%% file: main.tex
\documentclass[10pt,twocolumn,letterpaper]{article}
\usepackage[accsupp]{axessibility} 

\usepackage[pagenumbers]{cvpr} 

\input{preamble}

\definecolor{cvprblue}{rgb}{0.21,0.49,0.74}
\usepackage[pagebackref,breaklinks,colorlinks,citecolor=cvprblue]{hyperref}

\newcommand{\tablestyle}[2]{\setlength{\tabcolsep}{#1}\renewcommand{\arraystretch}{#2}\centering\footnotesize}

\def\confName{CVPR}
\def\confYear{2024}

\begin{document}

\title{
DiffusionMTL: Learning Multi-Task Denoising Diffusion Model  \\  from Partially Annotated Data}

\author{Hanrong Ye and Dan Xu\textsuperscript{\Letter}\\
Department of Computer Science and Engineering, HKUST\\
Clear Water Bay, Kowloon, Hong Kong\\
{\tt\small \{hyeae, danxu\}@cse.ust.hk}
}

\maketitle

\begin{abstract}
Recently, there has been an increased interest in the practical problem of learning multiple dense scene understanding tasks from partially annotated data, where each training sample is only labeled for a subset of the tasks. The missing of task labels in training leads to low-quality and noisy predictions, as can be observed from state-of-the-art methods. To tackle this issue, we reformulate the partially-labeled multi-task dense prediction as a pixel-level denoising problem, and propose a novel multi-task denoising diffusion framework coined as DiffusionMTL. It designs a joint diffusion and denoising paradigm to model a potential noisy distribution in the task prediction or feature maps and generate rectified outputs for different tasks. To exploit multi-task consistency in denoising, we further introduce a Multi-Task Conditioning strategy, which can implicitly utilize the complementary nature of the tasks to help learn the unlabeled tasks, leading to an improvement in the denoising performance of the different tasks. Extensive quantitative and qualitative experiments demonstrate that the proposed multi-task denoising diffusion model can significantly improve multi-task prediction maps, and outperform the state-of-the-art methods on three challenging multi-task benchmarks, under two different partial-labeling evaluation settings. The code is available at \url{https://prismformore.github.io/diffusionmtl/}.
\footnote{The paper is accepted by CVPR 2024.}
\end{abstract}

\section{Introduction}
\begin{figure}[t]
    \centering 
     \includegraphics[width=1.1\linewidth]{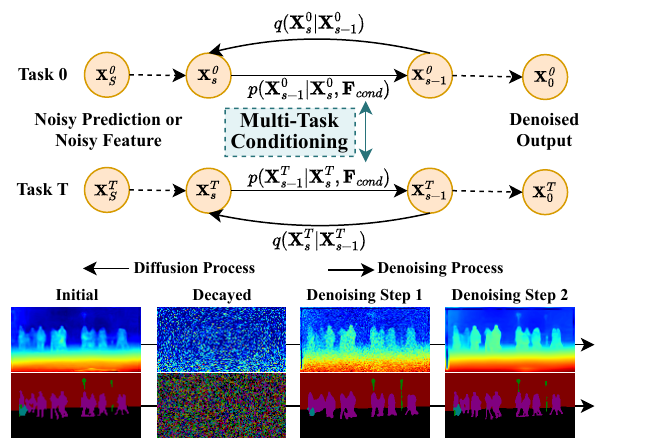}
     \vspace{-20pt}
     \caption{Motivative illustration of the proposed DiffusionMTL for multi-task partially supervised dense prediction. The model denoise the manually decayed multi-task prediction or feature maps~(denoted as $\{\mathbf{X}^0_S,...,\mathbf{X}^T_S\}$, $T, S$ are the numbers of tasks and steps separately) in a step by step manner, and obtain the denoised outputs $\{\mathbf{X}^0_0,...,\mathbf{X}^T_0\}$. The denoising process is guided by the designed multi-task condition feature $\mF_{cond}$. 
     }
     \vspace{-20pt}
     \label{fig:motivation}
\end{figure}

Multi-task learning for dense scene understanding~\cite{astmt,zamir2020robust,kokkinos2017ubernet,padnet} is an important research topic that has recently gained a lot of attention from computer vision researchers.
It aims at jointly learning multiple scene-related dense prediction tasks, including semantic segmentation, surface normal estimation, depth estimation, etc.
This multi-task learning problem has dual superiority over traditional single-task learning. On the one hand, multi-task models are naturally more efficient than single-task models with similar structures because different tasks can share some network modules. On the other hand,  different tasks are able to help each other and improve overall performance by sharing information through cross-task consistency~\cite{mtlsurvey}. 
However, annotating a real-world multi-task learning dataset at the pixel level is a daunting task.
As an alternative, collecting data annotated for different tasks and using them to train a multi-task model is a much more feasible approach.
This motivates recent work~\cite{weihongmtpsl} that defines an important new problem known as ``Multi-Task Partially Supervised Learning (MTPSL)", where each training sample contains labels for a subset of the tasks, rather than all tasks.
As there is a lack of multi-task labels for each training sample, the partially supervised multi-task learning problem is more challenging compared to the fully supervised multi-task learning problem.
To handle this problem, previous state-of-the-art models~\cite{weihongmtpsl} focus on improving label efficiency by enforcing cross-task consistency. They train an additional network to construct a joint feature space for each task pair, which helps improve the multi-task optimization process and demonstrates promising multi-task performance under MTPSL.
Despite their success in improving model performance, the sparsity of training labels in MTPSL still inevitably leads to noisy prediction maps which can be observed from previous state-of-the-art models, as shown in Figure~\ref{fig:qual_compare}.   
Therefore, there is a need for a new methodology to effectively denoise the noisy multi-task dense predictions to improve the multi-task prediction quality.

To address the above-mentioned problem, we propose a novel multi-task denoising diffusion framework that can effectively remove noise from the dense predictions and rectify multi-task prediction maps.
We formulate the multi-task dense prediction problem as a joint pixel-level denoising and generation process, and propose a new multi-task model coined as ``DiffusionMTL''. 
DiffusionMTL learns to denoise noisy multi-task predictions with the help of diffusion models~\cite{ho2020ddpm}, which are particularly effective in recovering data distribution from noisy input.
It jointly performs the diffusion and denoising processes to discover potential noisy distributions of the multi-task prediction maps, and learns to rectify the prediction maps. 
We further present two distinct diffusion mechanisms: Prediction Diffusion and Feature Diffusion. Prediction Diffusion learns to remove noise from the multi-task prediction maps, while Feature Diffusion learns to refine the multi-task feature maps.
Unlike typical diffusion models used for image synthesis~\cite{dhariwal2021diffusionbeat}, our denoising network is designed to achieve two objectives simultaneously. Firstly, it must reverse the Markovian noise diffusion process, \ie, remove the manually added noise from the input maps. Secondly, it is encouraged to generate higher-quality multi-task predictions from the noisy input, thereby improving overall multi-task prediction performance. 
Furthermore, to exploit multi-task consistency in the denoising process, we design a Multi-Task Conditioning mechanism for our DiffusionMTL model. This mechanism utilizes the prediction maps generated by the decoders of all the tasks to effectively facilitate the denoising process of a target task. 
Meanwhile, the outputs of the unlabeled tasks also receive supervision signals from the ground-truth labels of other tasks, allowing our DiffusionMTL to not only enhance the denoising performance of the labeled tasks but also facilitate the learning of unlabeled tasks. 

To evaluate the effectiveness of our approach for multi-task partially supervised learning, we have conducted extensive experiments on three challenging partially-annotated multi-task datasets, namely PASCAL, NYUD, and Cityscapes. Both quantitative and qualitative results demonstrate the effectiveness of the proposed DiffusionMTL model, and show that DiffusionMTL significantly outperforms the current state-of-the-art method by a large margin, using the same backbone and fewer model parameters.

\par 
In summary, the contribution of this paper is threefold:
\begin{itemize}
    \item We propose the first multi-task denoising diffusion framework for the partially labeled multi-task dense prediction problem. The innovative framework reformulates multi-task dense prediction as a joint pixel-level diffusion and denoising process, which empowers us to generate rectified higher-quality multi-task predictions.
    \item We develop a novel Multi-Task Denoising Diffusion Network specifically designed to address the issue of noise in initial prediction maps. An effective Multi-Task Conditioning mechanism is designed in our diffusion model to enhance the denoising performance. We further devise two effective diffusion mechanisms, namely Prediction Diffusion and Feature Diffusion, for refining task signals in prediction and feature spaces separately. 
    \item Extensive experiments have been conducted on three prevalent partial-labeling multi-task benchmarks under two different settings, which clearly validate the effectiveness of our proposal. Our method demonstrates significant performance improvements compared to the previous state-of-the-art methods. 
\end{itemize}

\begin{figure*}[t]
    \centering
    \vspace{-20pt}
     \includegraphics[width=1\linewidth]{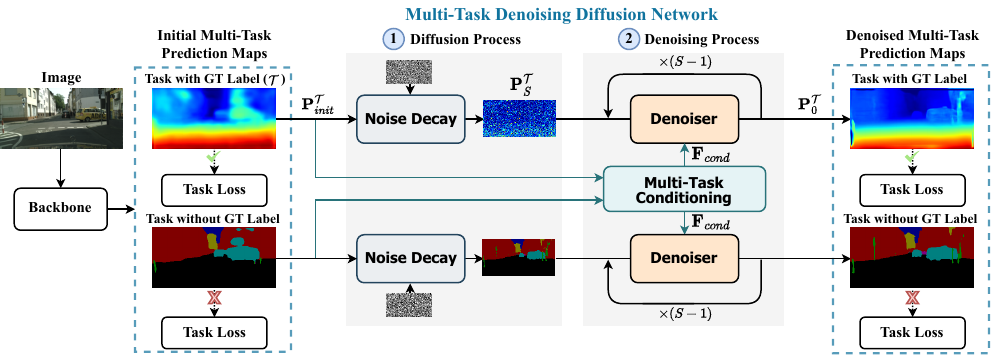}
     \vspace{-20pt}
     \caption{Illustration of the proposed DiffusionMTL~(Prediction Diffusion) framework for the MTPSL setting. DiffusionMTL first uses an initial backbone model for producing starter prediction maps for all tasks. 
     To denoise the initial prediction maps and generate rectified maps, we propose a \textbf{Multi-Task Denoising Diffusion Network}~(MTDNet). 
     MTDNet involves a diffusion process and a denoising process. 
     During training, the initial prediction map of the labeled target task $\mathcal{T}$ is gradually degraded by applying noise, resulting in the noisy prediction map~$\mP_S^\mathcal{T}$. Then, we utilize a Multi-Task Conditioned Denoiser~(referred to as the ``Denoiser") to denoise $\mP_S^\mathcal{T}$ iteratively over $S$ steps, resulting in a clean prediction map $\mP_0^\mathcal{T}$ that is supervised by the ground-truth label.
     For better learning of unlabeled tasks, 
     we propose a \textbf{Multi-Task Conditioning} mechanism in the denoising process to stimulate information sharing across different tasks.
     During inference, the diffusion and denoising processes are applied to all tasks to produce denoised multi-task prediction maps.
     }
     \label{fig:framework}
     \vspace{-10pt}
\end{figure*}

\begin{figure}[t]
    \centering
     \hspace*{-18pt}
     \includegraphics[width=1.2\linewidth]{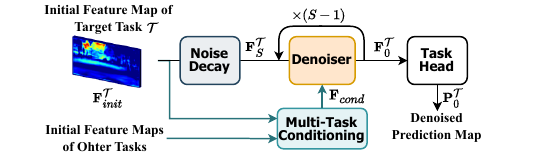}
     \vspace{-20pt}
     \caption{Illustration of the proposed DiffusionMTL~(Feature Diffusion), which conducts noise decay and denoising on initial feature maps $\mF_{init}^\mathcal{T}$. The denoised feature maps $\mF_{0}^\mathcal{T}$ are projected to the final prediction map $\mP_{0}^\mathcal{T}$ with a task head after the denoising.
     }
     \label{fig:fea_diffusion}
     \vspace{-15pt}
\end{figure}

\section{Related Work}
\noindent \textbf{Multi-Task Dense Scene Understanding with Partially Annotated Data}
Multi-task learning (MTL) for dense scene understanding has been widely studied in recent years~\cite{mtlsurvey,astmt,kokkinos2017ubernet,crossstitch,liu2019MTAN,zhang2021survey,zamir2018taskonomy,kanakis2023composite,liang2022m,chen2022mod,hoyer2021three}. 
By learning several tasks together, MTL enhances the computational efficiency of both training and inference compared to single-task models while achieving better performance~\cite{mti,atrc,padnet}. 
To improve the performance of multi-task learning, some researchers have focused on improving the optimization process of MTL by designing loss functions~\cite{weihongmtpsl,yang_contrastive,liu2022auto,zamir2020robust,kendall2018multi} and manipulating gradients~\cite{liu2021conflict,gradientsign,gradnorm,wang2020gradient,yu2020gradient}, while other researchers work on designing powerful multi-task model architectures~\cite{taskprompter2023,bachmann2022multimae,xu2023multi,ye2023taskexpert,xu2022mtformer,invpt2022,ye2023invpt++,li2022Universal,nddr,gao2020mtl,zhang2021automtl}. 
It is worth noting that the aforementioned methods are mainly designed for fully-supervised multi-task learning, where the labels of all tasks are assumed to be given for each training image. However, in real-world scenarios, it is not always feasible to obtain labels for all tasks, and we may have data with only some of the tasks available. 
To address this issue, a new problem named Multi-Task Partially Supervised Learning (MTPSL) has been defined by~\cite{weihongmtpsl}. In MTPSL, the training samples are only partially annotated for the tasks, which poses new challenges to multi-task learning due to the sparsity of labels in the training data. To meet the challenge, XTC~\cite{weihongmtpsl} has been proposed to better leverage partial annotations by improving label efficiency. It maps the label spaces of different tasks into one joint feature space and utilizes cross-task consistency to learn tasks without labels for each training sample. Although this approach has shown promising results, it still inevitably suffers from noisy predictions because the model is under-trained with a limited number of ground-truth labels.
To directly tackle the noisy prediction problem, our proposal takes a distinct approach by designing a novel multi-task denoising framework to improve the quality of multi-task prediction maps.

\begin{figure*}[t]
    \centering
     \vspace{-20pt}
     \includegraphics[width=1\linewidth]{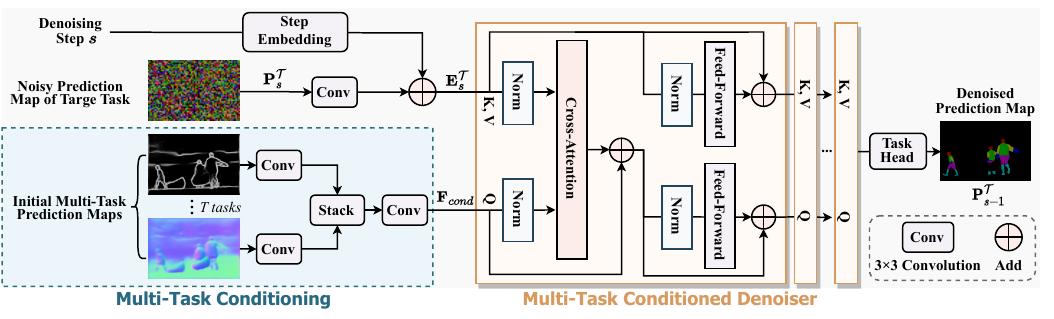}
     \vspace{-20pt}
     \caption{Pipeline of a single step $s$ in the denoising process of DiffusionMTL (Prediction Diffusion). 
     \textbf{Multi-Task Conditioning}:
     The initial prediction maps for all tasks are projected to task-specific features and then stacked. The stacked features are then processed with a $3\times3$ convolution to reduce the channel dimension, resulting in a Multi-Task Condition Feature $\mF_{cond}$ which is shared across all tasks. 
     \textbf{Multi-Task Conditioned Denoiser}:
     The denoiser consists of several cross-attention transformer blocks, which learn to denoise input conditioned on $\mF_{cond}$.
     For its input, we perform a $3\times3$ convolution on the noisy prediction map $\mP_s^\mathcal{T}$ and combine the output with the step embedding, obtaining a task embedding $\mE_{s}^{\mathcal{T}}$.
     The denoiser takes $\mF_{cond}$ as query input and $\mE_{s}$ as key and value inputs. 
     We use a task-specific head to obtain the denoised prediction map $\mP_{s-1}^\mathcal{T}$, which is the input of the next denoising step $s-1$.
}
    \vspace{-10pt}
     \label{fig:denoiser}
\end{figure*}

\noindent \textbf{Diffusion Models}
Diffusion models~\cite{sohl2015nonequilibrium,ho2020ddpm} are a class of generative models that have been widely used for image synthesis tasks, and have achieved state-of-the-art performance on several benchmarks~\cite{ho2020ddpm,kingma2021variational,dhariwal2021diffusionbeat,rombach2022latentdiff,Peebles2022DiT}. However, adapting diffusion models for multi-task dense prediction is not straightforward. Although some attempts have been made to apply diffusion models to single-task deterministic problems including image classification~\cite{han2022card}, segmentation~\cite{baranchuk2021labeleff,amit2021segdiff,chen2022generalist,gu2022diffusioninst,ye2023seggen} and detection~\cite{chen2022diffusiondet}, they are not suitable for multi-task dense scene understanding. 
In this paper, we propose a novel multi-task diffusion model that can denoise noisy prediction maps for multiple tasks and obtain finer results under the multi-task partially-supervised setting. 
Our approach represents an exploratory advancement in diffusion models and has the potential to inspire the design of future diffusion models for deterministic tasks. 

\section{The Proposed DiffusionMTL Approach}
In this section, we will introduce the details of our proposed multi-task denoising diffusion framework, DiffusionMTL, as illustrated in Fig.~\ref{fig:framework}. DiffusionMTL has two steps: (i) First, an initial backbone model generates preliminary prediction maps for multiple dense scene understanding tasks. (ii) Second, a proposed Multi-Task Denoising Diffusion Network (MTDNet) takes in the noisy initial multi-task prediction maps and produces refined prediction results. 
These two parts are trained together in an end-to-end manner with partially annotated data. 

\subsection{Initial Backbone Model}
\label{sec:init_model}
We adopt a classic encoder-decoder structure for the multi-task dense prediction~\cite{padnet,mti,invpt2022}. The initial backbone model utilizes a task-shared encoder $f_{enc}$, which accepts an input image $\mI \in \mathbb{R}^{H \times W \times 3}$~(where $H$ and $W$ represent height and width, respectively) and projects it to obtain a multi-channel backbone feature map $\mF_{backbone} \in \mathbb{R}^{H' \times W' \times C}$. The backbone feature map has a height of $H'$, a width of $W'$, and $C$ channels. It is shared by all the tasks.
And then, to generate task-specific feature maps for $T$ tasks, we adopt a series of task-specific decoders $\{f_{dec}^1, f_{dec}^2,..., f_{dec}^T\}$ with identical network structures and different network parameters. The generated initial task feature maps from the decoders are notated as $\{\mF_{init}^1,\mF_{init}^2,...,\mF_{init}^T\}$. 
For the $t$-th task, we compute the task-specific initial feature map $\mF_{init}^t$ as:
\begin{equation}
    \mF_{init}^t = f_{dec}^t ( f_{enc}(\mI) ).
\end{equation}
To compute an initial dense prediction map $\mP_{init}^t$ for the $t$-th task, we apply a task-specific~$1\times 1$ convolution~$f_{pred}^t$ on the corresponding task feature map $\mF_{init}^t$:
\begin{equation}
    \begin{aligned}
    \mP_{init}^t = f_{pred}^t  (\mF_{init}^t).
    \end{aligned}
\end{equation}
\noindent In this way, we obtain the $T$ initial  prediction maps of all $T$ tasks. The initial prediction maps are noisy as we can observe from Fig.~\ref{fig:framework}. 
We aim to rectify the noisy multi-task prediction maps with the following MTDNet.

\subsection{Multi-Task Denoising Diffusion Network}
In this paper, we put forward a novel diffusion model, named Multi-Task Denoising Diffusion Network~(MTDNet) for denoising the aforementioned noisy prediction maps. 
To achieve this goal, we design two orthogonal diffusion mechanisms in our unified MTDNet, focusing on different signal domains: 
\textbf{(i) Prediction Diffusion and (ii) Feature Diffusion}.
These mechanisms differ in terms of the signal space in which the diffusion model is applied.
Feature Diffusion refines the task-specific features within a high-dimensional latent space, while Prediction Diffusion directly improves the initial task predictions in the output space.
Feature Diffusion facilitates a comprehensive improvement of high-level visual information within an expanded latent space, while Prediction Diffusion demonstrates effective denoising capabilities along with better computational efficiency.
More details about their difference will be provided when describing the different components of MTDNet. As shown in Fig.~\ref{fig:framework} for Prediction Diffusion and Fig.~\ref{fig:fea_diffusion} for Feature Diffusion, we follow the DDPM paradigm~\cite{ho2020ddpm}, which is separated into 2 processes: a diffusion process and a denoising process.
During the diffusion process, we incrementally degrade the information in the initial prediction maps by applying noise with a Markovian chain. 
In the denoising process, we introduce a novel denoising network that is trained to generate clean prediction maps from the degraded ones in an iterative manner. 
Without loss of generality, we elaborate on the one-label setting of the multi-task partially supervised learning, where we assume that the current training sample has label for only one task (task $\mathcal{T}$). We name this labeled task as the ``target task".

\subsubsection{Diffusion Process}
\label{sec:diff}

In the diffusion process~(or ``forward process'')~\cite{ho2020ddpm}, we construct a fixed Markov chain with a total length of $S$ steps.
For the target task $\mathcal{T}$, 
Gaussian noise is gradually applied to the initial map $\mX_{init}^\mathcal{T}$ in a step-by-step manner. Here $\mX_{init}^\mathcal{T}$ is the initial prediction map $\mP_{init}^\mathcal{T}$ (in Prediction Diffusion) or initial task feature map $\mF_{init}^\mathcal{T}$ (in Feature Diffusion). 
Suppose the decayed map at denoising step $s$ of target task $\mathcal{T}$ is $\mX_s^\mathcal{T}$, the diffusion process $q$ can be formulated as:
\begin{equation}
    q(\mX_s^\mathcal{T}|\mX_{init}^\mathcal{T}) = \mathcal{N}(\mX_{s}^\mathcal{T}| \sqrt{\bar\alpha_s} \mX_{init}^\mathcal{T}, (1-\bar\alpha_s ) \mI),
\end{equation}
where $\{\bar\alpha_s, s\in 1,2,...,S \}$ are hyperparameters. In practice, we could directly compute the final decayed map.
Through mathematical derivation, the decayed prediction map of target task $\mathcal{T}$ at the final step $S$ can be formulated as 
$\mX_{S}^\mathcal{T}=\sqrt{\bar\alpha_S} \mX_{init}^\mathcal{T} +\sqrt{1-\bar\alpha_S }\epsilon$, where $\epsilon \sim \mathcal{N}(0,\mathbf{I})$.  For more theoretical details please refer to~\cite{ho2020ddpm}.

\makeatletter
\renewcommand{\ALG@name}{Pseudocode}
\makeatother

\begin{algorithm}[t]
\small
\caption{DiffusionMTL under one-label setting}
\begin{algorithmic}[1]
\Function{DiffusionMTL}{$\text{version}, \text{mode}, \mI, \mL, \mathcal{T}, S, T$}

\State \textbf{Input:} version $\in$ \{`Prediction Diffusion', `Feature Diffusion'\}, mode $\in$ \{`train', `infer'\}, input image $\mI$, label map $\mL$, target task $\mathcal{T}$, diffusion steps $S$, number of tasks $T$
\State \textbf{Output:} Training loss or denoised prediction map $\mP^\mathcal{T}_0$
\State $\mF_{backbone} \gets f_{enc}(\mI)$
\For{$t \gets 1,2,\dots,T$}
\State $\mF_{init}^t \gets f_{dec}^t(\mF_{backbone})$
\Comment{Compute initial task features}
\State $\mP_{init}^t \gets f_{pred}^t(\mF_{dec}^t)$
\Comment{Compute initial prediction maps}
\EndFor
\If{version $=$ `Prediction Diffusion'} 
    \State $\mX_{init}^\mathcal{T} \gets \mP_{init}^\mathcal{T}$ 
    \State $\mF_{cond} \gets f_{cond}(\mP_{init}^1, \mP_{init}^2, \dots, \mP_{init}^T)$
\ElsIf{version $=$ `Feature Diffusion'}
    \State $\mX_{init}^\mathcal{T} \gets \mF_{init}^\mathcal{T}$ 
    \State $\mF_{cond} \gets f_{cond}(\mF_{init}^1, \mF_{init}^2, \dots, \mF_{init}^T)$
\EndIf
\State $\epsilon \sim \mathcal{N}(0, \mI)$
\Comment{Sample noise}
\State $\mX_{S}^\mathcal{T} \gets \sqrt{\bar\alpha_S} \mX_{init}^\mathcal{T} + \sqrt{1-\bar\alpha_S} \epsilon$
\Comment{Diffusion process}
\For{$s \gets S, S-1, \dots, 1$}
\State $\mX^\mathcal{T}_{s-1} \gets \mathrm{Denoiser}(\mX^\mathcal{T}_{s}, s, \mF_{cond})$
\Comment{Denoising step}
\EndFor
\If{version $=$ `Prediction Diffusion'} 
    \State  $\mP_{0}^\mathcal{T} \gets \mX_{0}^\mathcal{T}$ 
\ElsIf{version $=$ `Feature Diffusion'}
    \State $\mP_{0}^\mathcal{T} \gets f_{head}^{\mathcal{T}}(\mX_{0}^\mathcal{T})$ \Comment{Final task head}
\EndIf
\If{mode $=$ `train'} \State \Return $\text{compute\_loss}(\mP^\mathcal{T}_0, \mL)$ \Comment{Compute loss with available label of target task} \ElsIf{mode $=$ `infer'}
\State \Return $\mP^\mathcal{T}_0$
\Comment{Output denoised prediction map}
\EndIf
\EndFunction
\end{algorithmic}
\label{alg:diffusionmtl}
\end{algorithm}

\subsubsection{Denoising Process}
\label{sec:reverse}

As the core component of our MTDNet, the denoising process involves designing a Multi-Task Conditioned Denoiser, referred to as ``Denoiser'', to denoise the noisy multi-task prediction maps or feature maps. 
Specifically, given the decayed noisy map of target task $\mX_S^\mathcal{T}$ from the diffusion process, 
Denoiser generates $\mX_{S-1}^\mathcal{T}$, $\mX_{S-2}^\mathcal{T}$,..., $\mX_{0}^\mathcal{T}$ in an iterative manner.
In the following, we will first introduce a novel Multi-Task Conditioning strategy, and then describe how Denoiser computes the denoised map in each denoising step.
Fig.~\ref{fig:denoiser} illustrates the computation pipeline for a single denoising step of Prediction Diffusion. 

\noindent\textbf{Multi-Task Conditioning}
To help denoise the prediction or feature maps of the labeled tasks, as well as enable learning unlabeled tasks under a partially annotated setting, we propose a Multi-Task Conditioning strategy in the denoising process.
It first obtains a ``multi-task condition feature''~(denoted as $\mF_{cond}$) from the initial maps of all the tasks. $\mF_{cond}$ captures the joint multi-task information, which is later used to condition the denoising network.
To obtain $\mF_{cond}$, we first project the initial multi-task maps to feature space via a $3\times 3$ convolution and obtain task-specific features for all $T$ tasks. 
Once we have obtained task-specific features for all $T$ tasks, we combine them along the channel dimension. The resulting tensor is then subjected to a $3\times3$ convolutional layer, which reduces the channel dimension to $C$ and then flattens the spatial dimension, resulting in a vector known as the multi-task condition feature $\mF_{cond}$. We refer to this computational process as $f_{cond}$.

\noindent\textbf{Multi-Task Conditioned Denoiser}
We illustrate the structure of denoiser in Fig.~\ref{fig:denoiser}. 
The denoiser is a series of cross-attention transformer blocks, taking in the noisy map $\mX_s^\mathcal{T}$, the denoising step $s$, and the multi-task condition feature $\mF_{cond}$ as input, and generates the denoised map $\mX_{s-1}^\mathcal{T}$. $\mX_{s-1}^\mathcal{T}$ is used as input for the next step in an iterative manner.
Specifically, we start by projecting the noisy map of the target task to a $C$-channel task embedding via a $3\times 3$ convolution and flatten its spatial dimension. Then we embed the denoising step $s$ using a typical sinusoidal embedding module~\cite{dhariwal2021diffusionbeat}. 
We call this process ``Step Embedding''. 
We add the step embedding to the task embedding.
The resulting task embedding $\mE^{\mathcal{T}}_s$ assumes the role of key and value tensors~($\mK, \mV$) in the subsequent transformer blocks, and $\mF_{cond}$ is supplied to the transformer blocks as the query $\mQ$:
\begin{equation}
    \mQ \leftarrow \mF_{cond},~~\mK\leftarrow \mE^{\mathcal{T}}_s,~~ \mV\leftarrow\mE^{\mathcal{T}}_s.
\end{equation}
The transformer blocks receive $\mQ$, $\mK$, and $\mV$ as input. Each block comprises linear normalization, cross-attention, and feed-forward networks as shown in Fig.~\ref{fig:denoiser}. For a more comprehensive understanding of the details of transformer, please consult~\cite{transformer}. Here, the cross-attention transformer blocks absorb information from the task embedding $\mE^{\mathcal{T}}_s$ guided by the multi-task conditioning feature $\mF_{cond}$.

The output procedure of a denoising step is different in Prediction Diffusion and Feature Diffusion.
In Prediction Diffusion, the output of the transformer blocks is reshaped to a spatial map and projected to prediction map $\mP_{s-1}^{\mathcal{T}}$ using a task-specific head which consists of several convolutional layers with ReLU.
In Feature Diffusion, the output of transformer blocks is projected to a feature map $\mF_{s-1}^{\mathcal{T}}$, which serves as the output of this step (\ie $\mX_{s-1}^{\mathcal{T}}$).
More implementation details can be found in Sec.~\ref{sec:imple_details}.
We can formulate each step in the denoising process as:
\begin{equation}
    \begin{aligned}
    \mX^\mathcal{T}_{s-1} &=  \mathrm{Denoiser}(\mX^\mathcal{T}_{s}, s, \mF_{cond}),\\
    s&\in S, S-1,..., 1.
    \end{aligned}
\end{equation}

For the final output after $S$ denoising steps, Prediction Diffusion directly generates the denoised prediction map of target task $\mP_0^\mathcal{T}$, which is used to compute task-specific loss supervised by the available ground-truth label.
In Feature Diffusion, we need a final task-specific head $f_{head}^\mathcal{T}$ to project the denoised feature maps $\mF_0^\mathcal{T}$ to the final prediction map $\mP_0^\mathcal{T}$.
We present the detailed training and inference pipelines of DiffusionMTL in Pseudocode~\ref{alg:diffusionmtl}.

\subsection{Model Optimization}
The whole DiffusionMTL model can be trained under the MTPSL setting in an end-to-end manner. Specifically, for each training sample, we apply task-specific losses on both the initial prediction maps as well as the final denoised prediction maps of the tasks with labels. For the unlabeled tasks, there are no ground-truth supervision signals, but the task-specific decoders are able to be implicitly trained via the proposed Multi-Task Conditioning. More details about losses are introduced in the supplemental materials.

\section{Experiments}

\subsection{Experimental Setup}
\noindent \textbf{Datasets and Tasks}
Following the pioneering work in multi-task partially supervised learning~\cite{weihongmtpsl}, we adopt three prevalent multi-task datasets with dense annotations, \ie PASCAL~\cite{everingham2010pascal}, NYUD~\cite{silberman2012indoor}, and Cityscapes~\cite{Cordts2016Cityscapes}.
\textbf{PASCAL} is a comprehensive dataset providing images of both indoor and outdoor scenes. There are  4,998 training images and 5,105 testing images, with labels of semantic segmentation, human parsing, and object boundary detection. Additionally, \cite{astmt} generates pseudo labels for surface normals estimation and saliency detection. 
\textbf{NYUD} (or NYUD-v2) provides images of indoor scenes as well as dense annotations for 13-class semantic segmentation and depth estimation. The images are resized to 288$\times$384. The surface normals can be generated from depth. The training set contains 795 images, while the testing set contains 654 images.
\textbf{Cityscapes} captures street scenes of different cities with fine pixel-level annotations. Following~\cite{weihongmtpsl,liu2019MTAN}, we use 7-class semantic segmentation and monocular depth estimation tasks in the experiments. The images are resized to 128$\times$256. There are 2,975 training images and 500 validation images.

\noindent \textbf{Task Metrics}
We adopt the same metrics for different tasks as previous work~\cite{weihongmtpsl}. We use the mean Intersection over Union (mIoU) to evaluate the performance of the semantic segmentation (Semseg) and human parsing (Parsing) tasks, while the absolute error (absErr) is used for evaluating the monocular depth estimation task (Depth). For the surface normal estimation task (Normal), we use the mean error of angles  (mErr) as the evaluation metric, while for the saliency detection task (Saliency), we use the maximal F-measure (maxF). The object boundary detection task (Boundary) is evaluated using the optimal-dataset-scale F-measure (odsF). To quantify the overall multi-task performance relative to the single-task baseline, we calculate the mean relative difference across all tasks, denoted as Multi-task Performance (MTL Perf $\Delta_m$)~\cite{astmt}.

\noindent \textbf{MTPSL Evaluation Settings}
There are two evaluation settings for multi-task partially supervised learning~\cite{weihongmtpsl}:
(i) \textbf{one-label setting}, where each training image has the ground-truth label of only one task.
(ii) \textbf{random-label setting}, where the number of labeled tasks for each image is random.
We use exactly the same image-task label mappings as~\cite{weihongmtpsl} for a strictly fair comparison.

\begin{figure}[t]
    \centering
     \includegraphics[width=\linewidth]{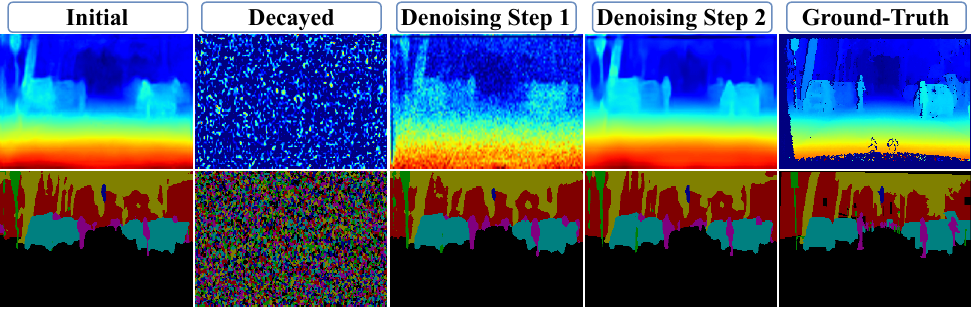}
     \vspace{-20pt}
     \caption{Visualization of the prediction maps at different processes on Cityscapes. 
     Our DiffusionMTL effectively denoises the noisy prediction maps of both tasks.
     }
     \label{fig:vis_reverse_three}
     \vspace{-10pt}
\end{figure}

\begin{table}[!t]
    \centering
\tablestyle{2pt}{1.1}
\scriptsize
\begin{tabular}{clcccccccccc}
        \toprule
         \multirow{2}{*}{\textbf{\# labels}} &  \multicolumn{1}{c}{\multirow{2}{*}{\textbf{Method}}} & 
         \textbf{Semseg}  &  \textbf{Depth} &  \textbf{MTL Perf} \\
        && mIoU $\mathbf{\uparrow}$ & absErr $\mathbf{\downarrow}$ &  $\Delta_m$ $\mathbf{\uparrow}$  \\
        \midrule
        \multirow{7}{*}{one / random} & Single-Task & 75.82 & 0.0125 & -\\
        \cline{2-5}
        & MTL Baseline & 73.19 & 0.0168 & -18.81\%  \\
        & SS~\cite{weihongmtpsl} & 71.67 & 0.0178 & -\\
        & XTC~\cite{weihongmtpsl} & 74.90 & 0.0161 & -\\
        & XTC*~\cite{weihongmtpsl}  & 73.36 & 0.0158 & -14.74\% \\
        & \textbf{DiffusionMTL (Prediction)} & 74.90 & 0.0131 & -2.79\% \\
        &\textbf{DiffusionMTL (Feature)} & \textbf{75.67} & \textbf{0.0130} & \textbf{-2.13\%} \\
	\bottomrule
 \end{tabular}%
   \vspace{-8pt}
    \caption{Comparison with SOTAs on Cityscapes. The proposed DiffusionMTL demonstrates superior performance on both tasks. One-label setting is equivalent to the random-label setting on Cityscapes. ``*'' denotes re-implemented results.}
\label{tab:cs}
  \vspace{-15pt}
\end{table}%

\begin{figure*}[t]
\vspace{-30pt}
\begin{minipage}[t]{0.4\textwidth}
  \centering
    \includegraphics[width=1\linewidth]{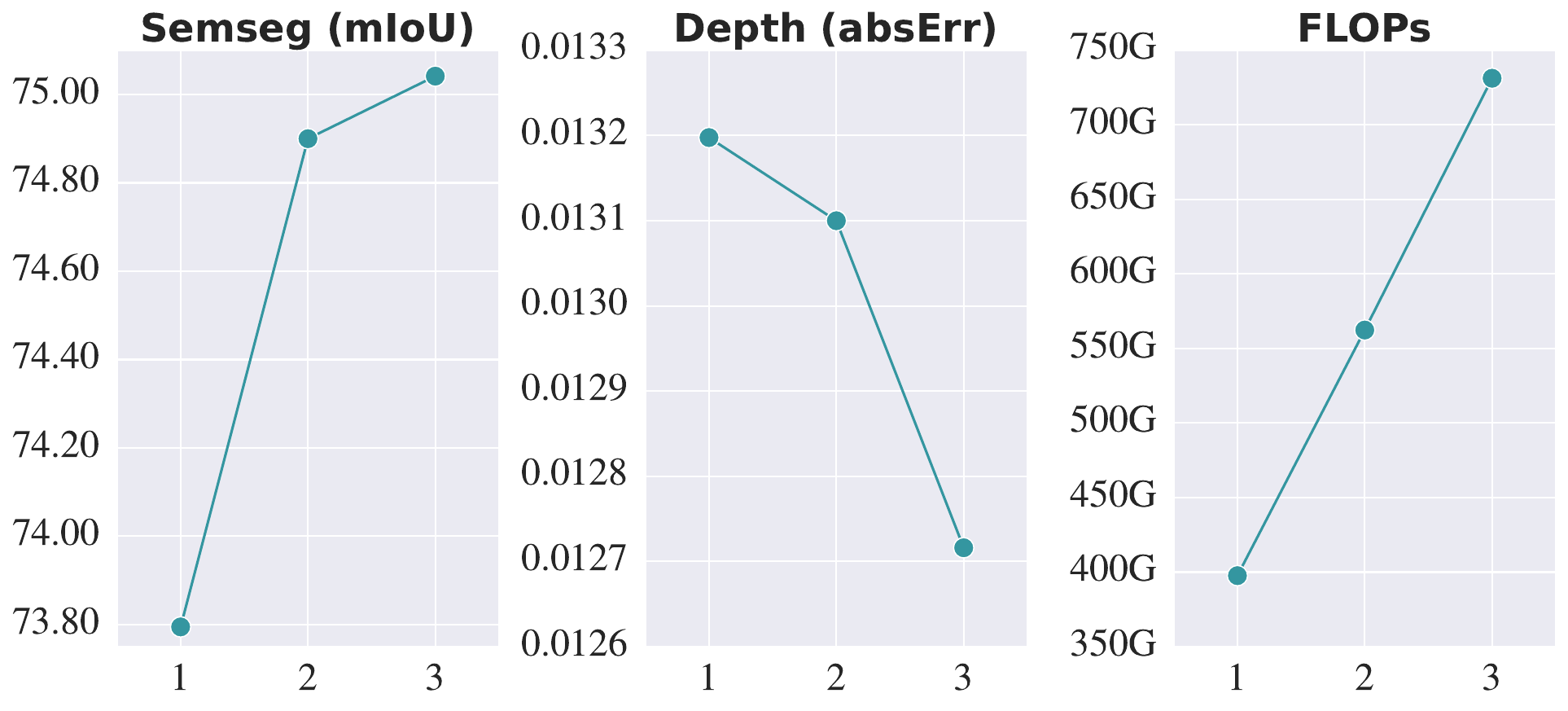}
     \vspace{-20pt}
  \captionof{figure}{Study of training DiffusionMTL with varying numbers of diffusion steps on Cityscapes.  Adding diffusion steps increases the performance and FLOPs.
  }
    \label{fig:diff_steps}
\end{minipage}\hfill
\begin{minipage}[t]{0.58\textwidth}
  \centering
    \vspace{-92pt}
    \resizebox{1.\linewidth}{!}{
    \tablestyle{1pt}{1.1}
    \scriptsize
    \begin{tabular}{clcccccccccc}
        \toprule
       \multicolumn{1}{c}{\multirow{2}{*}{\textbf{Backbone}}} & \multicolumn{1}{c}{\multirow{2}{*}{\textbf{Method}}} &  \multirow{2}{*}{\textbf{\#Params}} & \multirow{2}{*}{\textbf{FLOPS}}  & \textbf{Semseg}  & \textbf{Parsing}  & \textbf{Saliency} & \textbf{Normal} & \textbf{Boundary}&  \textbf{MTL Perf}  \\
       & &&& mIoU $\mathbf{\uparrow}$  & mIoU $\mathbf{\uparrow}$
      & maxF $\mathbf{\uparrow}$ & mErr $\mathbf{\downarrow}$ & odsF $\mathbf{\uparrow}$ &  $\Delta_m$ $\mathbf{\uparrow}$\\
        \midrule
        \multirow{4}{*}{R18} & \textbf{DiffusionMTL (Feature)}  & 133M & 676G & 57.78 & \textbf{58.98} & \textbf{77.82} & \textbf{16.11} & \textbf{64.50} & \textbf{+3.65\%} \\
         & \textbf{DiffusionMTL (Prediction)}  & 133M & 628G & \textbf{59.43} & 56.79 & 77.57 & 16.20 & 64.00 & +3.23\%  \\
        &---w/o Diffusion & 133M & 628G & 57.77 & 56.39 & 77.44 & 17.34 & 60.60 & +0.11\% \\  
        &---w/o Multi-Task Cond & 125M & 558G & 55.95 & 55.90 & 77.22 & 17.87 & 61.50 & -1.33\%\\
        \midrule
         \multirow{4}{*}{R50} &\textbf{DiffusionMTL (Feature)} & 159M & 742G & 58.78 & 	\textbf{61.91} & 77.07 & \textbf{16.49} & \textbf{66.20} &\textbf{ 0.77\%} \\
         & \textbf{DiffusionMTL (Prediction)} & 159M & 694G & \textbf{60.92} & 59.94 & \textbf{77.58} & 17.31 & 63.80 & -0.66\% \\
         &---w/o Diffusion  & 159M & 694G & 58.10 & 58.69 & 76.64 & 17.50 & 62.80 & -2.86\% \\
         &---w/o Multi-Task Cond & 150M & 625G & 57.29 & 59.37 & 76.90 & 17.74 & 63.70 & -2.91\% \\
        \bottomrule
    \end{tabular}%
    }
    \vspace{-5pt}
  \captionof{table}{Ablation study on PASCAL. ``w/o Diffusion'' indicates replacing the diffusion model with an iterative refinement model using an identical network structure. ``w/o Multi-Task Cond'' means removing Multi-Task  Conditioning. }
    \label{tab:ablation}
\end{minipage}
\end{figure*}

\begin{table*}[!t]
\scriptsize
\centering
 \vspace{-5pt}
\tablestyle{2pt}{1.1}
\scriptsize
    \begin{tabular}{cl|cccccccc|cccccccc}
        \toprule
        \multirow{3}{*}{\textbf{\# labels}} &  \multicolumn{1}{c|}{\multirow{3}{*}{\textbf{Method}}} & \multicolumn{8}{c|}{\textbf{PASCAL}} & \multicolumn{4}{c} {\textbf{NYUD}} \\
        \cline{3-14}
        && \multirow{2}{*}{\textbf{\#Params}} & \multirow{2}{*}{\textbf{FLOPS}} & \textbf{Semseg}  & \textbf{Parsing}  & \textbf{Saliency} & \textbf{Normal} & \textbf{Boundary}&  \textbf{MTL Perf} & \textbf{Semseg}  &  \textbf{Depth}  & \textbf{Normal} &  \textbf{MTL Perf} \\
        &&&& mIoU $\mathbf{\uparrow}$  & mIoU $\mathbf{\uparrow}$
      & maxF $\mathbf{\uparrow}$ & mErr $\mathbf{\downarrow}$ & odsF $\mathbf{\uparrow}$ &$\Delta_m$ $\mathbf{\uparrow}$ & mIoU $\mathbf{\uparrow}$ & absErr $\mathbf{\downarrow}$ & mErr $\mathbf{\downarrow}$  &$\Delta_m$ $\mathbf{\uparrow}$ \\
        \midrule
        \multirow{7}{*}{one}
        & Single-Task Baseline & 219M & 817G & 50.34 & 59.05	 & 77.43 & 16.59 & 64.40 & - & 45.28 & 0.4802 & 25.93 & -\\
        \cline{2-14}
        & MTL Baseline & 157M & 608G & 49.71	& 56.00 & 74.50	& 16.85	& 62.80 & -2.85\% & 43.92 & 0.5138 & 26.44 & -3.99\% \\
        & SS~\cite{weihongmtpsl}  & -& - & 45.00 & 54.00 & 61.70 & 16.90 & 62.40 & - & 27.52 &  0.6499 &  33.58 & - \\
        & XTC~\cite{weihongmtpsl}  & -& - & 49.50 & 55.80 & 61.70 & 17.00 & \textbf{65.10} & - & 30.36 & 0.6088 & 32.08 & -\\
        & XTC*~\cite{weihongmtpsl} & 173M & 608G & 55.08 & 56.72 & 77.06 & 16.93 & 63.70 & +0.37\% & 43.97 & 0.5140 & 26.30 & -3.79\% \\
        & \textbf{DiffusionMTL (Prediction)} & 133M & 628G  & \textbf{59.43} & 56.79 & 77.57 & 16.20 & 64.00 & +3.23\% & \textbf{44.97} & 0.5137 & 26.17 & -2.86\% \\
        &\textbf{DiffusionMTL (Feature)}  & 133M & 676G & 57.78 & \textbf{58.98} & \textbf{77.82} & \textbf{16.11} & 64.50 & \textbf{+3.65\%} & 44.47 & \textbf{0.5059} & \textbf{25.84} & \textbf{-2.27}\% \\
        \midrule
        \multirow{7}{*}{random}
        & Single-Task & 219M & 817G & 51.51 & 57.90 & 80.30 & 15.24 & 67.80 & - & 48.25 & 0.4792 & 24.65 & -\\
        \cline{2-14}
        & MTL Baseline & 157M & 608G & 62.23 & 55.88 & 78.67 & 15.47 & 66.70 & +2.44\% & 45.93 & 0.4839 & 25.53 & -3.12\% \\
        & SS~\cite{weihongmtpsl}  & -& - & 59.00 & 55.80 & 64.00 & 15.90  &  66.90 & - & 29.50 & 0.6224 & 33.31 & - \\
        & XTC~\cite{weihongmtpsl} &-&-& 59.00 & 55.60 & 64.00 & 15.90 & \textbf{67.80} &- & 34.26 & 0.5787 & 31.06 & -\\
        & XTC*~\cite{weihongmtpsl} & 173M& 608G & 62.44 & 55.81 & 78.56 & 15.45 & 66.80 & +2.52\% & 46.03 & 0.4811 & 25.97 & -3.44\% \\
        & \textbf{DiffusionMTL (Prediction)} & 133M & 628G & \textbf{63.68} & 55.84 & 79.87 & 15.38 & 66.80 & +3.44\% & \textbf{47.44} & 0.4803 & 25.26 & -1.45\% \\
        &\textbf{DiffusionMTL (Feature)} & 133M & 676G & 62.55 & \textbf{56.84} & \textbf{80.44} & \textbf{14.85} & 67.10 & \textbf{+4.27\%} & 46.82 & \textbf{0.4743} & \textbf{24.75} & \textbf{-0.77\%} \\
        \bottomrule
    \end{tabular}%
    \vspace{-8pt}
    \caption{Quantitative multi-task performance comparison with the state-of-the-arts (SOTAs) on PASCAL and NYUD. All models use a ResNet-18 as the backbone. `one' means each training image has only one labeled task, while `random' means each training image has a random number of labeled tasks. The proposed DiffusionMTL, including both Prediction Diffusion and Feature Diffusion, achieves significantly better performance while using fewer model parameters. ``*'' denotes our re-implemented results.}
    \label{tab:pascal}
    \vspace{-15pt}
\end{table*}%

\noindent\textbf{Implementation Details}
\label{sec:imple_details}
For most models in the experiments, we use ResNet-18 as backbone with ImageNet pre-trained weights provided by PyTorch. We concatenate the feature maps of the four stages of ResNet-18 along the channel dimension and process them with a $3\times3$ convolution to reduce the number of channels to 512. 
For our DiffusionMTL, the initial multi-task backbone has a task-specific decoder for each task, using 2 residual convolutional blocks~\cite{resnet}, followed by a $1\times1$ convolution as prediction head. 
Each residual convolutional block contains two $3\times3$ convolutions with BN and ReLU.
The denoising network uses 4 task-shared cross-attention transformer blocks, which are followed by a task-specific head of four $3\times3$ CONV-ReLU layers and a $1\times1$ convolution for task prediction in Prediction Diffusion.
We use 2 steps in the diffusion process.
More implementation details can be found in the supplemental materials.

\subsection{Main Experiments}
\noindent\textbf{Declaration of Comparison Models}
\label{sec:model_declaration}
We consider several models for comparison to verify the effectiveness of our proposed DiffusionMTL framework:
(i)  ``MTL Baseline''  is the baseline model. It shares the same backbone as DiffusionMTL and utilizes a strong task-specific decoder for each task. It consists of 6 residual convolutional blocks, followed by a $1\times1$ convolution as prediction head.
(ii) ``SS'' and ``XTC''~\cite{weihongmtpsl} are pioneering state-of-the-art methods as introduced in related work. XTC is re-implemented on our MTL Baseline based on their official codes for fair comparison.  
(iii) ``Single-Task'' is a single-task learning version of the MTL Baseline. It contains a set of separate models where each model is trained to learn only one single task.

\begin{figure}[t]
    \centering
     \includegraphics[width=1\linewidth]{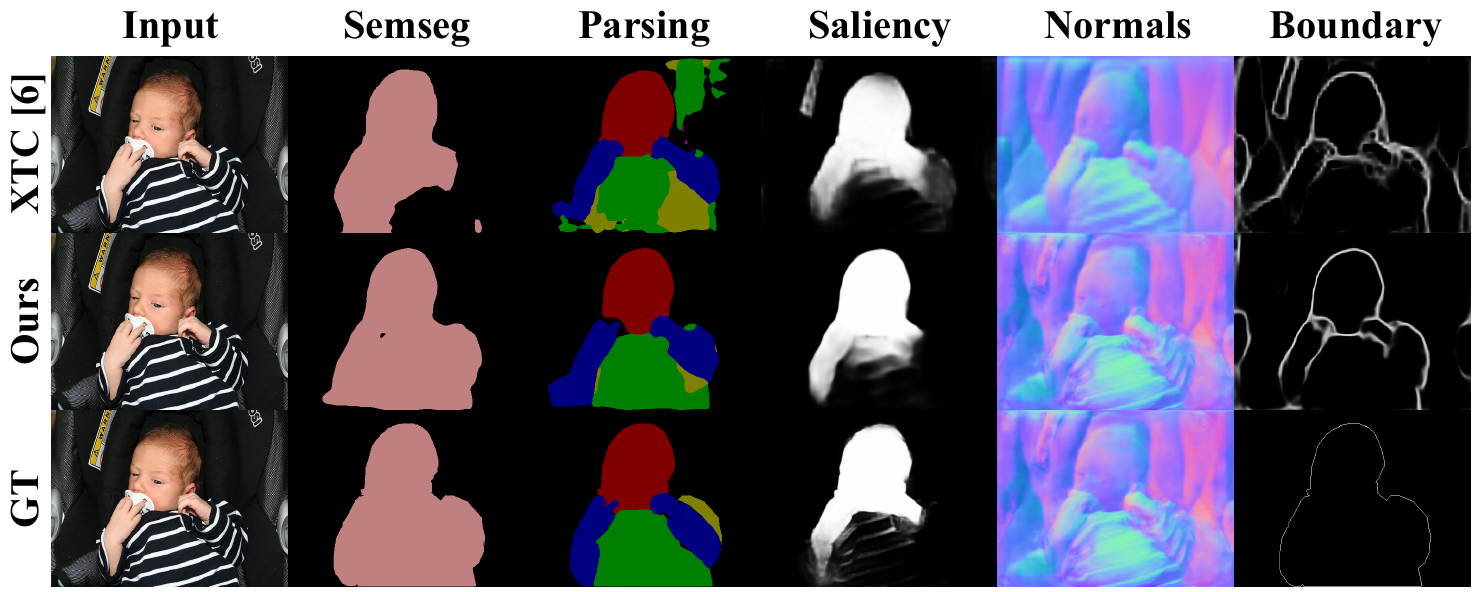}
     \vspace{-20pt}
     \caption{Qualitative comparison between the state-of-the-art XTC~\cite{weihongmtpsl} and our method on PASCAL under one-label setting. XTC suffers from the issue of noisy predictions. In contrast, our DiffusionMTL model learns to denoise the noisy prediction maps, resulting in significantly better multi-task prediction maps.}
     \label{fig:qual_compare}
     \vspace{-18pt}
\end{figure}

\begin{figure*}[ht]
    \centering
    \vspace{-30pt}
     \includegraphics[width=1\linewidth]{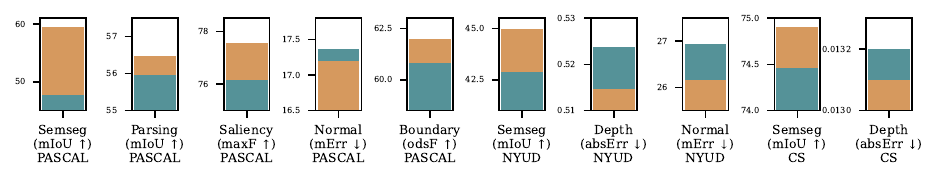}
     \vspace{-27pt}
     \caption{Performance of the initial multi-task predictions (\textcolor[HTML]{559298}{blue}) and the denoised predictions (\textcolor[HTML]{d6995e}{yellow}) of DiffusionMTL (Prediction) on PASCAL, NYUD, and Cityscapes under one-label setting. Our denoising network improves the prediction quality of all 10 tasks. }
     \label{fig:plot_gain}
     \vspace{-15pt}
\end{figure*}

\begin{figure}[t]
    \centering
     \includegraphics[width=1\linewidth]{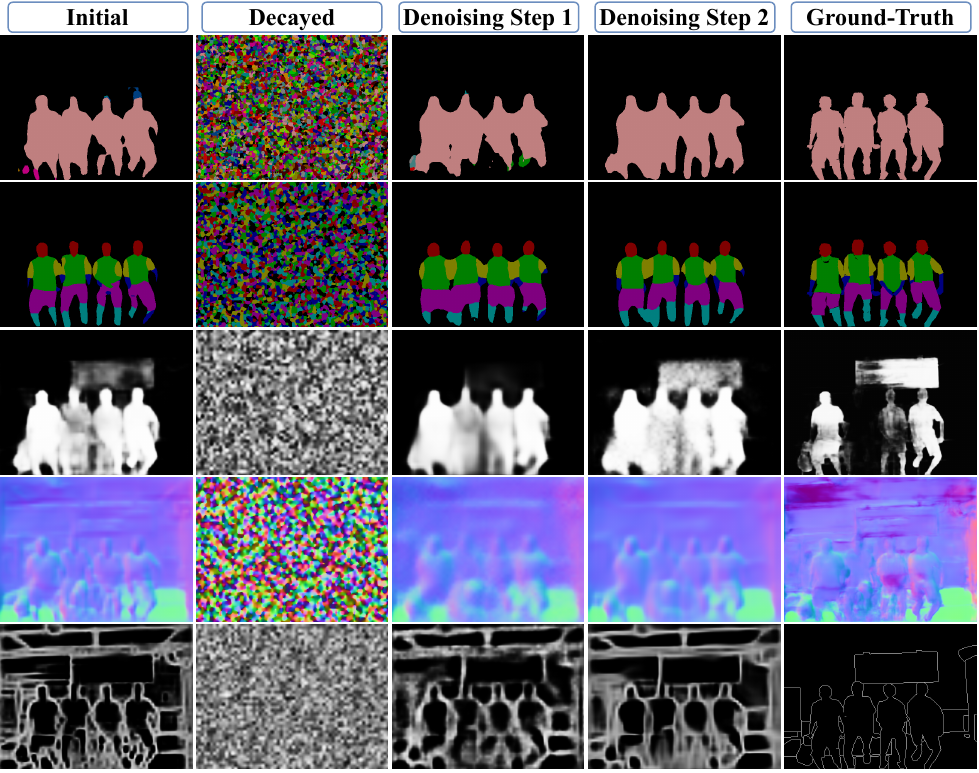}
     \vspace{-20pt}
     \caption{
     Visualization of the prediction maps at different processes on PASCAL.
     Our DiffusionMTL can denoise and rectify the noisy multi-task prediction maps.
     }
     \label{fig:vis_reverse}
     \vspace{-17pt}
\end{figure}

\noindent\textbf{Quantitative Comparison with SOTAs}
We compare the proposed DiffusionMTL with several strong competitors introduced in Sec.~\ref{sec:model_declaration} on three widely-used benchmarks, \ie~Cityscapes, PASCAL, and NYUD. We show the results of the three benchmarks in Table~\ref{tab:cs} and Table~\ref{tab:pascal}, respectively. 
As can be observed from the tables, our DiffusionMTL demonstrates significant improvements over the competing MTL Baseline and XTC~\cite{weihongmtpsl} on all three benchmarks. 
Specifically, under the challenging one-label setting on PASCAL, our DiffusionMTL (prediction) outperforms the MTL Baseline by +9.72 on Semseg and +3.07 on Saliency, while the multi-task performance $\Delta_m$ is also improved by +6.08\%. 
Compared with the state-of-the-art method XTC, our proposal achieves an improvement of +2.86\% in terms of the multi-task performance $\Delta_m$. We can observe consistent performance gains under the random-label setting.
Similarly, under the one-label setting of NYUD, our Feature Diffusion improves $\Delta_m$ by +1.52\% compared with XTC.
On Cityscapes, where the one-label setting is equivalent to the random-label setting, the multi-task performance $\Delta_m$ is improved by +11.95\% compared with the previous best XTC. For computational efficiency, as shown in Table~\ref{tab:pascal}, our model consumes only 133M network parameters, which is clearly less than 173M used by XTC~\cite{weihongmtpsl}.
These significant results can fully show the effectiveness of the proposed DiffusionMTL method, which substantially outperforms the competing methods on all benchmarks while using fewer model parameters.

\noindent\textbf{Qualitative Comparison with SOTAs}
To examine the quality of generated multi-task predictions by DiffusionMTL, we visualize the predictions by Prediction Diffusion trained under the one-label setting of PASCAL, and compare them with the outputs of SOTA model~\cite{weihongmtpsl} as well as ground-truth labels in Fig.~\ref{fig:qual_compare}. 
The images are randomly chosen from the testing set of PASCAL.
We observe that the generated multi-task predictions of the previous best method are noisy, which confirms our motivation to design a multi-task denoising framework for the multi-task partially supervised learning problem.
With the proposed DiffusionMTL, the prediction quality is significantly improved.

\subsection{Ablation Study}
We conduct comprehensive ablation experiments to evaluate the effectiveness of different components of DiffusionMTL for multi-task partially supervised learning and show the results on PASCAL one-label setting in Table~\ref{tab:ablation}.  

\noindent\textbf{Effectiveness of Multi-Task Denoising Diffusion Network}
To further confirm the effectiveness of the proposed multi-task denoising diffusion network (MTDNet), we replace it with an iterative refinement network using an identical network structure (\ie cross-attention transformer blocks in Prediction Diffusion). This variant is denoted as ``w/o Diffusion'' in  Table~\ref{tab:ablation}. 
MTDNet brings a significant multi-task performance improvement of +3.12~(ResNet-18) and +2.20~(ResNet-50) using the same computational costs, which clearly validates the effectiveness of the proposed multi-task denoising method. 
Moreover, we plot the performance metrics of initial predictions and final predictions in Fig.~\ref{fig:plot_gain}, which shows improvement brought by MTDNet on all 10 tasks of three benchmarks.

\noindent \textbf{Effectiveness of Multi-Task Conditioning}
To evaluate the efficacy of the multi-task conditioning strategy, we conduct ablation experiments by removing it from DiffusionMTL~(Prediction) and replacing cross-attention blocks with self-attention blocks. This variant is indicated as ``w/o Multi-Task Cond''.
Multi-task conditioning leads to significant improvement on all tasks, underscoring the unique importance of multi-task information sharing in the partially-labeled multi-task learning problem.

\noindent \textbf{Comparison of Feature Diffusion and Prediction Diffusion}
As observed in Table~\ref{tab:ablation}, both multi-task diffusion mechanisms show significant multi-task performance on different backbones. Prediction Diffusion is more computationally efficient due to the lower dimensions of the intermediate maps, whereas Feature Diffusion achieves higher performance on most tasks by capturing more visual information in the features.

\noindent \textbf{Qualitative Analysis of Denoising Effect}
In the denoising process, the model is designed to denoise the noisy multi-task prediction maps that are degraded by the diffusion process. To evaluate the denoising performance of our Prediction Diffusion, we provide visualizations of the multi-task prediction maps at different phases in Fig.~\ref{fig:vis_reverse_three} and Fig.~\ref{fig:vis_reverse}. Our model generates clean and accurate multi-task prediction maps from noisy inputs, which firmly indicates the effectiveness of our multi-task denoising framework.

\noindent \textbf{Influence of Diffusion Steps}
We plot the performance metrics against diffusion steps on Cityscapes in Fig.~\ref{fig:diff_steps}. 
Our observations reveal that utilizing two steps yields a notable improvement in performance compared to using just one step. Moreover, increasing the number of steps further enhances performance, albeit at a higher computational cost. Therefore, we set the default number of steps to two.

\section{Conclusion}
Our study aims to address the issue of noisy predictions in multi-task learning from partially annotated data. We propose a unified multi-task denoising diffusion framework that refines multi-task signals in the feature and prediction spaces separately. Additionally, we introduce an effective Multi-Task Conditioning strategy to enhance denoising performance and facilitate learning of unlabelled tasks through cross-task information sharing. Extensive experiments on three prevalent datasets validate our approach, which outperforms previous methods by a significant margin.

\noindent\textbf{Acknowledgement}
This research is partially supported by the Early Career Scheme of the Research Grants Council (RGC) of the Hong Kong SAR under grant No. 26202321.

{\small
\bibliographystyle{unsrt}
\bibliography{refers}
}

\clearpage
\appendix

\section{Supplemental Implementation Details}
\subsection{Additional Details about DiffusionMTL}
\textbf{Multi-Task Denoising Diffusion Network.}
This section provides additional details about the implementation of DiffusionMTL on different datasets. For our experiments, we set the default diffusion steps to 2 using a linear variance scheduler with a range from $10^{-3}$ to $10^{-2}$. All self-attention blocks in the Denoiser use a single head. 

\noindent \textbf{Loss functions.} 
For semantic segmentation, human parsing, saliency detection, and boundary detection, we use cross-entropy loss. For depth and surface normal estimation, we opt for L1 loss.
The multi-task loss balance weights are the same as those used in~\cite{weihongmtpsl}.

\subsection{Implementation Details on Different Datasets}
For all three partial-labeling benchmarks (PASCAL, NYUD, and Cityscapes), we use exactly the same image-task label mappings as those used in~\cite{weihongmtpsl}.

\noindent\textbf{PASCAL}
On PASCAL-Context~\cite{everingham2010pascal}~(abbreviated as ``PASCAL''), in the one-label setting, there are 1000, 999, 1000, 1000, 999 images separately labeled for semantic segmentation, human parsing, surface normal estimation, saliency detection, and boundary detection.
In the random-label setting, there are 450, 2553, 2480, 2445, and 2557 images labeled for semantic segmentation, human parsing, surface normal estimation, saliency detection, and boundary detection, respectively,
We pad the images to a resolution of $512\times512$. We use the Adam optimizer and a polynomial learning rate scheduler with a base learning rate of $2\times 10^{-5}$. All models are trained for 100 epochs with a batch size of 6.
We adopt the same data augmentations as in~\cite{mti}, which include random scaling, cropping, random horizontal flipping, and color jittering.

\noindent \textbf{NYUD}~\cite{silberman2012indoor}
In the one-label setting, 265 images are labeled for semantic segmentation, 265 images are labeled for monocular depth estimation, and 265 images are labeled for surface normal estimation. In the random-label setting, 392, 408, and 385 images are respectively labeled for these tasks.
The images are resized to a resolution of $288\times384$. We use the Adam optimizer and a polynomial learning rate scheduler with a base learning rate of $2\times 10^{-5}$. All models are trained for 200 epochs with a batch size of 4.
We adopt the same data augmentations as in~\cite{weihongmtpsl}, which include random cropping and random horizontal flipping.

\noindent \textbf{Cityscapes}~\cite{Cordts2016Cityscapes} 
As we only evaluate two tasks on the Cityscapes dataset, the one-label setting is equivalent to the random-label setting. The training split contains 1,487 labeled images for semantic segmentation and 1,488 labeled images for monocular depth estimation.
We adopt a learning rate of $10^{-4}$. All models are trained for 200 epochs with a batch size of 8. The images are resized to a resolution of $128\times256$.
We adopt the data augmentations in~\cite{weihongmtpsl}, which include random cropping and random horizontal flipping.

\section{Additional  Quantitative Study}
\subsection{Comparison with SOTA refinement methods.}
We conduct extensive experiments to compare our proposal with previous SOTA MTL refinement methods, including  MTI-Net~\cite{mti} and InvPT~\cite{invpt2022}, based on the ResNet-18 baseline under the one-label setting on PASCAL dataset. The results, presented in Table~\ref{tab:refine}, demonstrate the superior performance of DiffusionMTL across all tasks.

\begin{table}[t]
\huge
\centering
    \resizebox{1.\linewidth}{!}
    {
    \begin{tabular}{lccccccccccc}
        \toprule
         \multicolumn{1}{l}{\multirow{2}{*}{\textbf{Method}}} & \multirow{2}{*}{\textbf{\#Params}} & \multirow{2}{*}{\textbf{FLOPS}} & \textbf{Train}  & \textbf{Semseg}  & \textbf{Parsing}  & \textbf{Saliency} & \textbf{Normal} & \textbf{Boundary}&  \textbf{MTL Perf} \\
        && & \textbf{GPU Mem} & mIoU $\mathbf{\uparrow}$  & mIoU $\mathbf{\uparrow}$
      & maxF $\mathbf{\uparrow}$ & mErr $\mathbf{\downarrow}$ & odsF $\mathbf{\uparrow}$ &$\Delta_m$ $\mathbf{\uparrow}$\\
        \midrule
        MTL Baseline & 157M  & 608G & 6163M & 49.71 & 56.00& 	74.50	& 16.85	& 62.80 & -2.85\% \\
        XTC~\cite{weihongmtpsl} & 173M  & 608G & 6409M  & 55.08 & 56.72 & 77.06 & 16.93 & 63.70 & +0.37\%\\
        MTINet~\cite{mti} & 281M & 589G & 11533M & 54.32 & 57.73 & 77.12 & 16.41 & 64.20 & +1.21\% \\
        InvPT~\cite{invpt2022} & 141M & 1182G & 9993M   & 56.96 & 57.05 & 77.19 & 16.80 & 63.20 & +1.27\% \\
        \midrule
        DiffusionMTL (Prediction) & 133M  & 628G & 5703M  & \textbf{59.43} & 56.79 & 77.57 & 16.20 & 64.00 & +3.23\% \\
        DiffusionMTL (Feature) & 133M & 676G & 5811M & 57.78 & \textbf{58.98} & \textbf{77.82} & \textbf{16.11} & \textbf{64.50} & \textbf{+3.65\%} \\
        \bottomrule
    \end{tabular}%
        }
    \vspace{-10pt}
    \caption{One-label setting on PASCAL with ResNet-18 backbone.}
    \label{tab:refine}
    \vspace{-15pt}
\end{table}%

\subsection{Comparison under Fully-Annotated Setting}
Our method can be applied to fully-annotated benchmarks.
We conduct experiments on fully-annotated PASCAL dataset using ResNet-18 and show the results in Table~\ref{tab:fully_supervised}.
Our method demonstrates stronger performance compared to both the baseline as well as the state-of-the-art~(SOTA) method XTC~\cite{weihongmtpsl} and InvPT~\cite{invpt2022}.

\begin{table}[h]
\huge
\centering
    \resizebox{1.\linewidth}{!}
    {
    \begin{tabular}{lccccccccccc}
        
        \toprule
         \multicolumn{1}{l}{\multirow{2}{*}{\textbf{Method}}} & \multirow{2}{*}{\textbf{\#Params}} & \multirow{2}{*}{\textbf{FLOPS}}  & \textbf{Semseg}  & \textbf{Parsing}  & \textbf{Saliency} & \textbf{Normal} & \textbf{Boundary}&  \textbf{MTL Perf} \\
        &&  & mIoU $\mathbf{\uparrow}$  & mIoU $\mathbf{\uparrow}$
      & maxF $\mathbf{\uparrow}$ & mErr $\mathbf{\downarrow}$ & odsF $\mathbf{\uparrow}$ &$\Delta_m$ $\mathbf{\uparrow}$\\
        \midrule
        STL Baseline & 219M &  817G  &  52.56 & 62.21 & 82.75 & 14.12 & 68.90 & -\\
        \midrule
        MTL Baseline & 157M  & 608G  & 62.91  & 57.37  & 81.82  & 14.49  & 66.40 & +0.90\% \\
        XTC~\cite{weihongmtpsl} & 173M  & 608G & 63.29 & 57.93 & 82.09 & 14.48 & 66.50 & +1.34\% \\
        InvPT~\cite{invpt2022} & 141M & 1182G  & 64.38 & 59.49 & 83.52 & 14.75 & 66.80 & +2.31\% \\
        \midrule
        DiffusionMTL (Prediction) & 133M  & 628G & 64.31 & 58.68 & 83.07 & 14.44 & 67.10 & +2.44\%\\
        DiffusionMTL (Feature) & 133M & 676G & \textbf{64.62} & \textbf{60.14} & \textbf{83.99} & \textbf{14.17} & \textbf{67.80} & \textbf{+3.84}\% \\
        \bottomrule
    \end{tabular}%
        }
    \vspace{-10pt}
    \caption{Fully-annotated setting on PASCAL with ResNet-18.}
    \label{tab:fully_supervised}
    \vspace{-12pt}
\end{table}%

\subsection{Computation and Memory Cost Comparison.}
We have already shown the parameters and FLOPs comparison with the MTL baseline and XTC in Table 3 of our main paper. We further provide the training GPU memory in Table~\ref{tab:refine} of this document. Our method shows higher parameter/memory efficiency and comparable computational costs with significantly better performance. 

\section{Additional  Qualitative Study}
\subsection{Denoising Effectiveness of DiffusionMTL}
\begin{figure*}[h]
    \centering
    \includegraphics[width=0.90\linewidth]{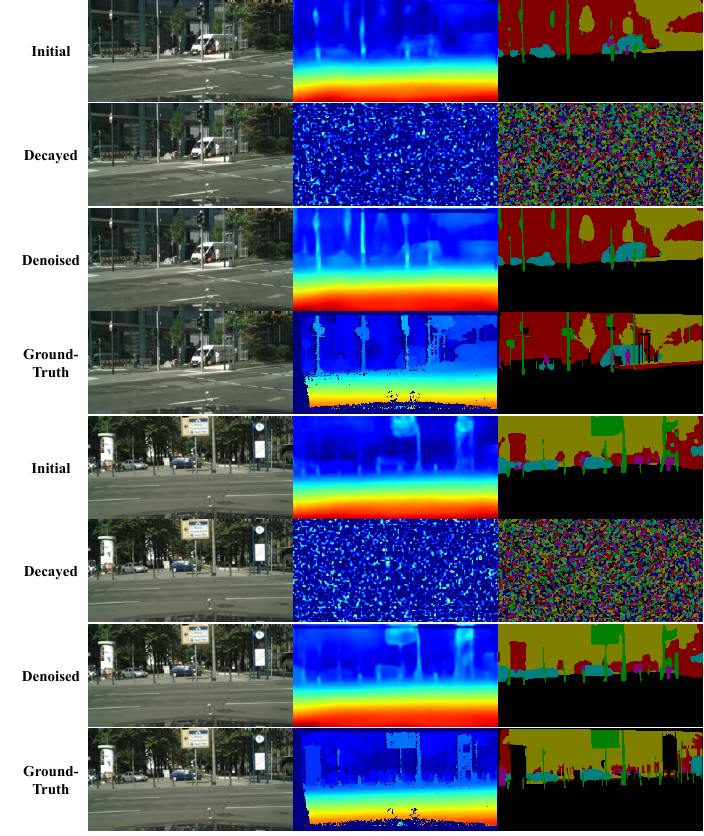}
     \vspace{-10pt}
     \caption{Qualitative comparison of the initial multi-task predictions, decayed predictions, our denoised results, and ground-truth labels on Cityscapes under one-label setting. Our DiffusionMTL is able to rectify noisy input and generate clean prediction maps. The model used in this comparison is trained on the Cityscapes dataset under the one-label MTPSL setting.}
     \label{fig:qualitative_compare_denoised}
\vspace{-10pt}
\end{figure*}

\begin{figure*}[!t]
    \centering
    \includegraphics[width=0.90\linewidth]{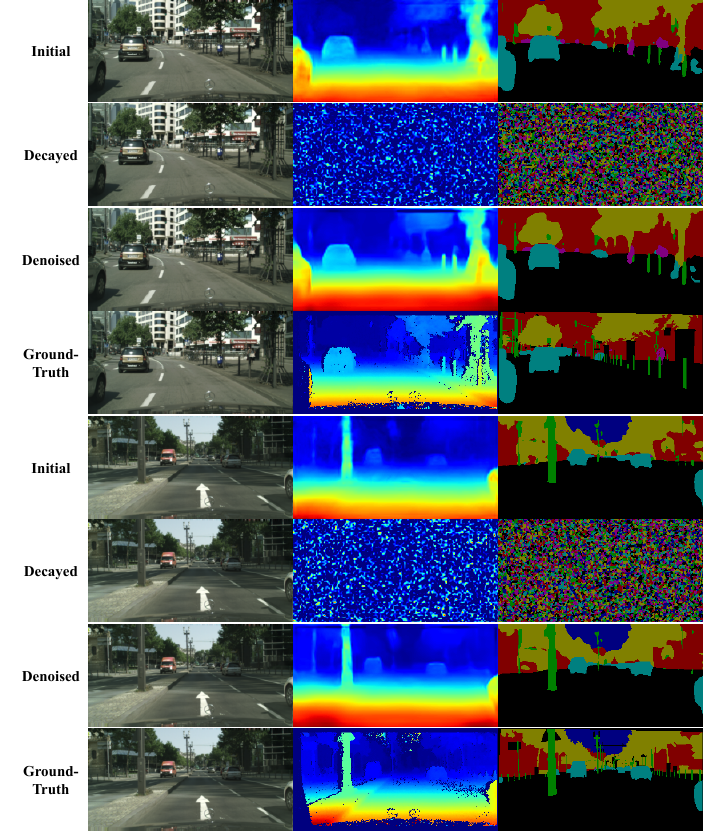}
     \vspace{-10pt}
     \caption{Qualitative comparison of the initial multi-task predictions, decayed predictions, our denoised results, and ground-truth labels on Cityscapes under one-label setting. Our DiffusionMTL is able to rectify noisy input and generate clean prediction maps. The model used in this comparison is trained on the Cityscapes dataset under the one-label MTPSL setting.}
     \label{fig:qualitative_compare_denoised_2}
\end{figure*}

To assess the denoising performance of our model, we visually examine the noisy multi-task prediction maps generated through the diffusion process, as well as the denoised outputs produced by Prediction Diffusion based on ResNet-18 on Cityscapes dataset under a one-label training setting. The obtained results are showcased in Fig.~\ref{fig:qualitative_compare_denoised} and Fig.~\ref{fig:qualitative_compare_denoised_2}. 
The effectiveness of our proposed DiffusionMTL is demonstrated by its ability to successfully denoise the noisy prediction maps, resulting in significantly improved multi-task predictions that align better with the ground-truth labels. These results serve as additional validation for our motivation behind designing a robust multi-task denoising diffusion framework, addressing the challenges inherent in the multi-task partially supervised learning problem. 

\subsection{Comparison with SOTA}
\begin{figure*}[!t]
    \centering
    \includegraphics[width=0.9\linewidth]{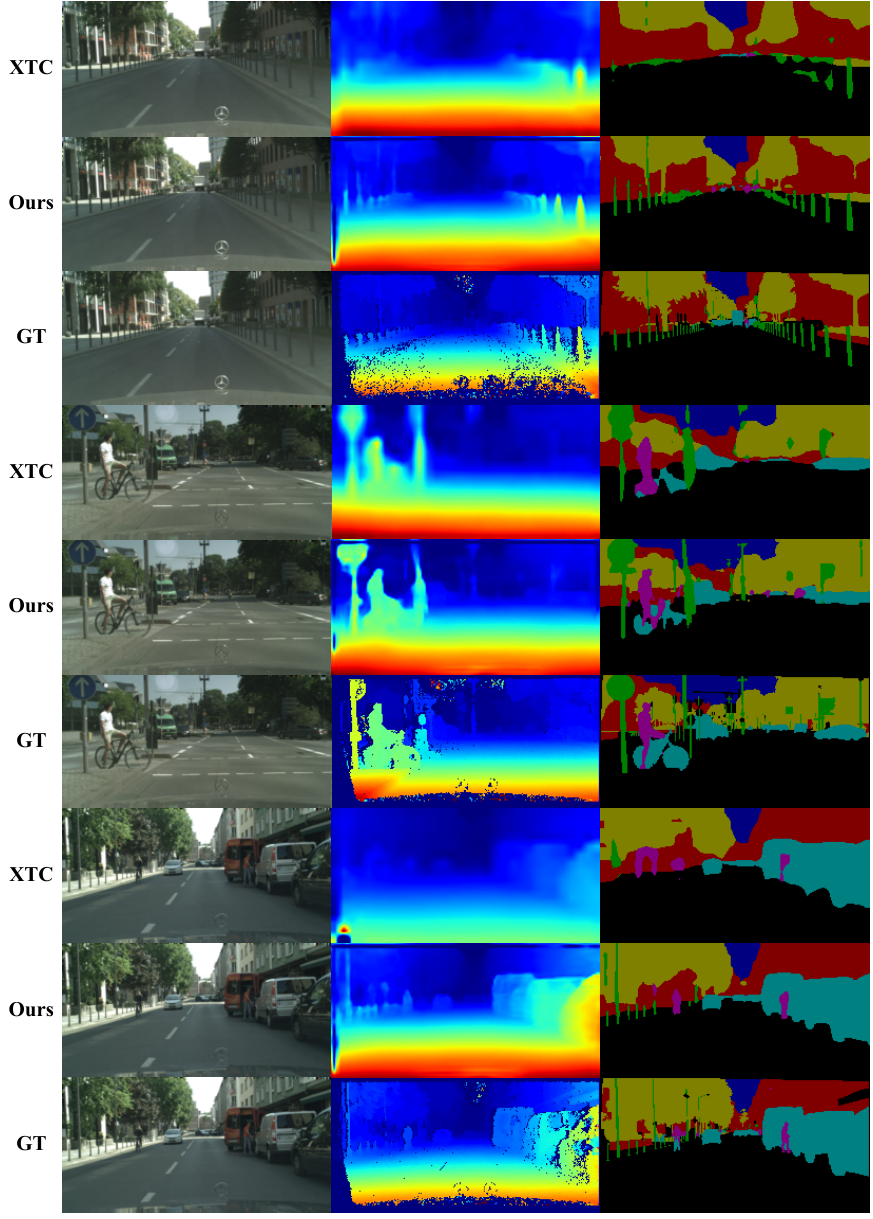}
     \vspace{-10pt}
     \caption{Qualitative comparison between our method and the state-of-the-art method (\ie XTC~\cite{weihongmtpsl}) for depth estimation and semantic segmentation tasks in Cityscapes dataset, using the same ResNet-18 backbone. Our DiffusionMTL approach outperforms the previous state-of-the-art method in producing superior prediction maps. Notably, each training sample is labeled for only one task.}
     \label{fig:qualitative_cs_sota}
\end{figure*}

\begin{figure*}[!t]
    \centering
    \vspace{-15pt}
    \includegraphics[width=0.9\linewidth]{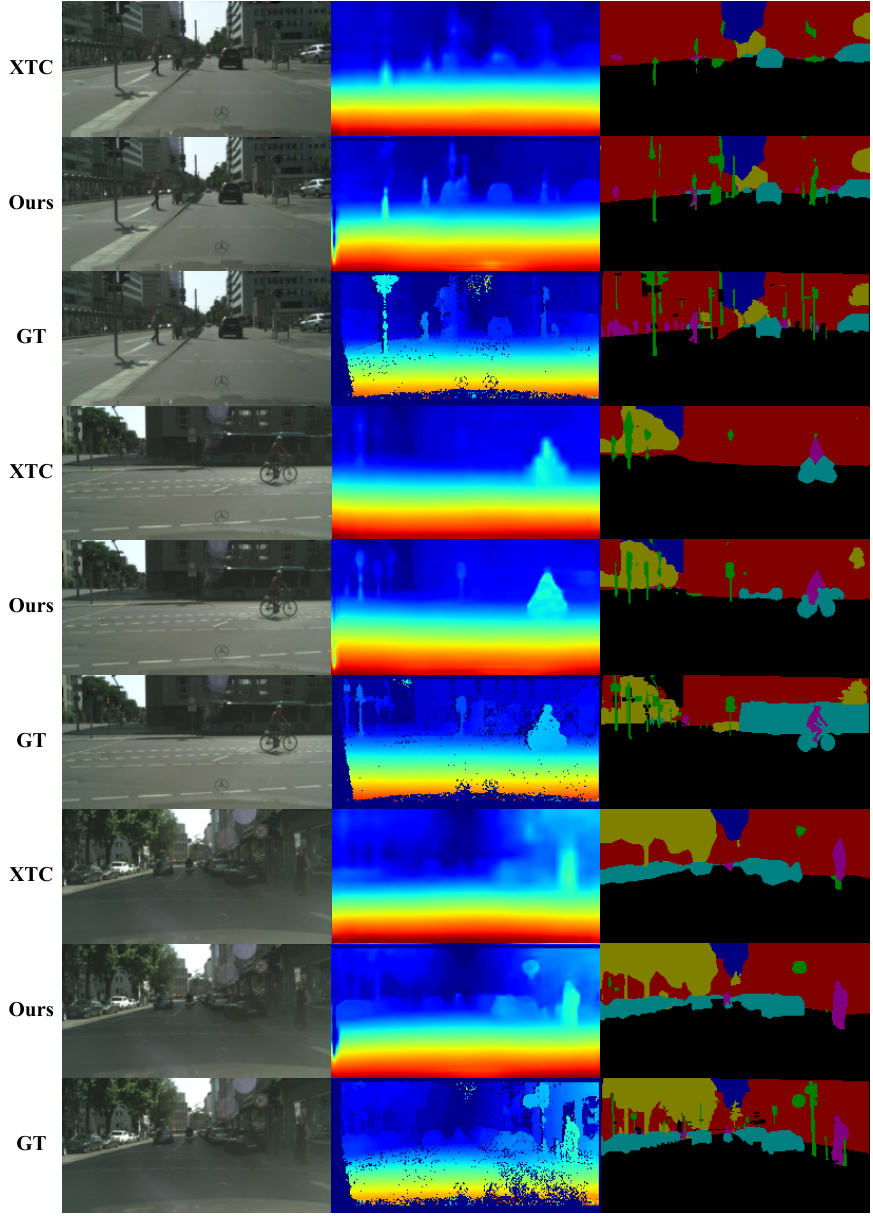}
     \vspace{-10pt}
     \caption{Qualitative comparison between our method and the state-of-the-art method (\ie XTC~\cite{weihongmtpsl}) for depth estimation and semantic segmentation tasks on the Cityscapes dataset, using the same ResNet-18 backbone. Our DiffusionMTL approach outperforms the previous state-of-the-art method in producing superior prediction maps. Notably, each training sample is labeled for only one task.}
     \label{fig:qualitative_cs_sota_1}
\end{figure*}

\begin{figure*}[!t]
    \centering
    \vspace{-15pt}
    \includegraphics[width=0.9\linewidth]{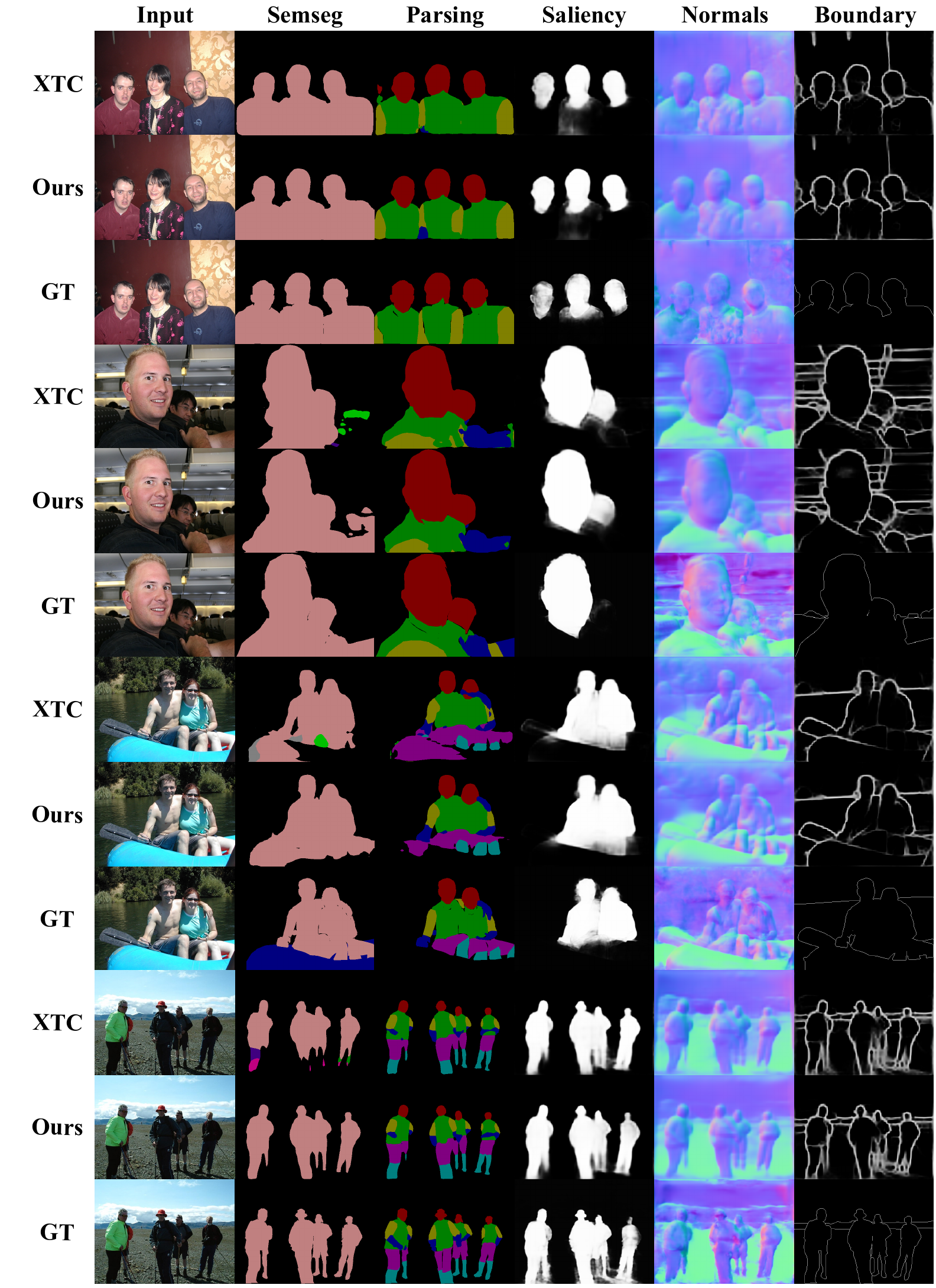}
     \vspace{-10pt}
     \caption{Qualitative comparison between our method and the state-of-the-art method (\ie XTC~\cite{weihongmtpsl}) in PASCAL dataset, using the same ResNet-18 backbone. Our DiffusionMTL approach outperforms the previous state-of-the-art method in producing superior prediction maps. Notably, each training sample is labeled for only one task.}
     \label{fig:qualitative_pascal_sota}
\end{figure*}

\begin{figure*}[!t]
    \centering
    \vspace{-15pt}
    \includegraphics[width=0.9\linewidth]{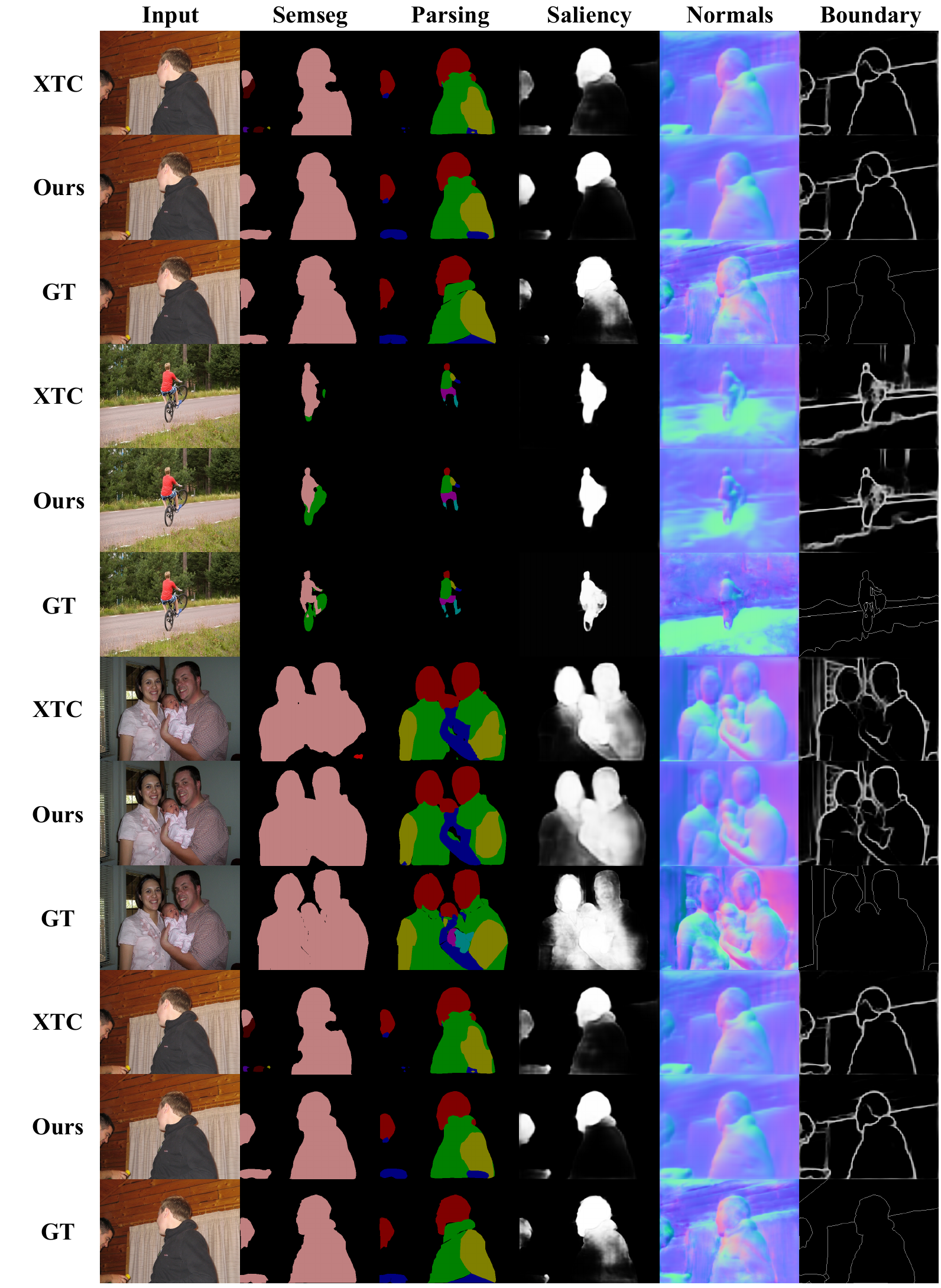}
     \vspace{-10pt}
     \caption{Qualitative comparison between our method and the state-of-the-art method (\ie XTC~\cite{weihongmtpsl}) in PASCAL dataset, using the same ResNet-18 backbone. Our DiffusionMTL approach outperforms the previous state-of-the-art method in producing superior prediction maps. Notably, each training sample is labeled for only one task.}
     \label{fig:qualitative_pascal_sota1}
\end{figure*}

In order to further demonstrate the performance advantage of DiffusionMTL, we present a set of randomly selected samples generated by our model and the previous state-of-the-art model (\ie, XTC~\cite{weihongmtpsl}) on Cityscapes in Fig.~\ref{fig:qualitative_cs_sota} and Fig.~\ref{fig:qualitative_cs_sota_1}. We further compare the results on PASCAL in Fig.~\ref{fig:qualitative_pascal_sota} and Fig.~\ref{fig:qualitative_pascal_sota1}. These models are trained under the same one-label multi-task partially supervised learning setting. The superiority of prediction maps generated by DiffusionMTL in terms of accuracy is evident on both datasets. This compelling evidence serves to further validate the effectiveness of our proposed denoising diffusion model.

\end{document}

%% file: preamble.tex
\usepackage[dvipsnames]{xcolor}

\usepackage{times}
\usepackage{epsfig}
\usepackage{graphicx}
\usepackage{amsmath}
\usepackage{amssymb}

\usepackage{booktabs}
\usepackage{url}
\usepackage{enumitem}
\usepackage{tikz}
\usepackage{comment}
\usepackage{color}
\usepackage{textcomp}  
\usepackage{amsfonts}
\usepackage{multirow}
\usepackage{cite}

\usepackage{algorithm}
\usepackage{algpseudocode}
\usepackage{listings}

\usepackage{float}
\usepackage{tabularx}
\usepackage{array}
\usepackage[misc]{ifsym}
\usepackage{caption}
\usepackage{lipsum}
\usepackage{wrapfig}
\usepackage{makecell} 
\input{math_commands.tex}

\usepackage{courier}

\usepackage{xcolor}
\usepackage{soul}
\definecolor{my_blue}{HTML}{296980}
\definecolor{my_red}{HTML}{ff5f2e}
\definecolor{my_yellow}{HTML}{FFF2CC}

\usepackage{gradient-text}

%% file: math_commands.tex
\usepackage{amsmath,amsfonts}
\def\bm#1{\mathbf{#1}}

\def\eqref#1{equation~\ref{#1}}

\def\1{\bm{1}}

\def\mE{{\bm{E}}}
\def\mF{{\bm{F}}}

\def\mI{{\bm{I}}}

\def\mK{{\bm{K}}}
\def\mL{{\bm{L}}}

\def\mP{{\bm{P}}}
\def\mQ{{\bm{Q}}}

\def\mV{{\bm{V}}}

\def\mX{{\bm{X}}}

\DeclareMathAlphabet{\mathsfit}{\encodingdefault}{\sfdefault}{m}{sl}
\SetMathAlphabet{\mathsfit}{bold}{\encodingdefault}{\sfdefault}{bx}{n}

